\documentclass[runningheads,a4paper]{llncs}

\usepackage{amssymb}
\usepackage{graphicx}
\usepackage{paralist}
\usepackage{enumitem}
\setitemize{noitemsep,topsep=0pt,parsep=0pt,partopsep=0pt}
\usepackage{url}
\urldef{\mailsa}\path|{martin.zaefferer,|
\urldef{\mailsb}\path|frederik.rehbach}@th-koeln.de|    
\newcommand{\keywords}[1]{\par\addvspace\baselineskip
\noindent\keywordname\enspace\ignorespaces#1}
  \usepackage{booktabs}
  \usepackage{paralist}
  \usepackage{relsize}
  \usepackage[utf8]{inputenc} 
  \usepackage{xspace}

\usepackage[linesnumbered,ruled,vlined]{algorithm2e}
\usepackage{url}
\usepackage{hyperref}
\hypersetup{hidelinks=true}

\usepackage{amstext}
\usepackage{amsmath}
\usepackage{bm}
\usepackage{cleveref}
  \makeatletter
\renewcommand\section{\@startsection{section}{1}{\z@}%
                       {-18\p@ \@plus -4\p@ \@minus -4\p@}%
                       {6\p@ \@plus 4\p@ \@minus 4\p@}%
                       {\normalfont\large\bfseries\boldmath
                        \rightskip=\z@ \@plus 8em\pretolerance=10000 }}
 \renewcommand\subsection{\@startsection{subsection}{2}{\z@}%
                       {-10\p@ \@plus -2\p@ \@minus -2\p@}%
                       {2\p@ \@plus 2\p@ \@minus 2\p@}%
                       {\normalfont\normalsize\bfseries\boldmath
                        \rightskip=\z@ \@plus 8em\pretolerance=10000 }}  \makeatother

\newcommand{\kf}{\ensuremath{{\text{k}}}\xspace}

\newcommand{\f}{\ensuremath{{\text{f}}}\xspace}
\newcommand{\fs}{\ensuremath{{\text{f}_\text{s}}}\xspace}
\newcommand{\fsc}{\ensuremath{{\text{f}_\text{sc}}}\xspace}

\newcommand{\hatmu}{\ensuremath{\hat{\mu}}\xspace}
\newcommand{\omegavec}{\ensuremath{\bm{\omega}}\xspace}

\newcommand{\vecx}{\ensuremath{\mathbf{x}}\xspace}
\newcommand{\vecy}{\ensuremath{\mathbf{y}}\xspace}
\newcommand{\veck}{\ensuremath{\mathbf{k}}\xspace}

\newcommand{\yvec}{\ensuremath{\mathbf{y}}\xspace}

\newcommand{\yvechs}{\ensuremath{\mathbf{\haty}_\text{s}}\xspace}
\newcommand{\evec}{\ensuremath{\mathbf{1}}\xspace} 
\newcommand{\epsvec}{\ensuremath{{\bm{\epsilon}}}\xspace}

\newcommand{\haty}{\ensuremath{\hat{y}}\xspace}
\newcommand{\RR}{\ensuremath{\mathbb{R}}\xspace}

\newcommand{\Xmat}{\ensuremath{\mathbf{X}}\xspace}
\newcommand{\Xs}{\ensuremath{{\mathbf{X}_\text{s}}}\xspace}

\newcommand{\Kmat}{\ensuremath{{\mathbf{K}}}\xspace}
\newcommand{\Kmats}{\ensuremath{{\mathbf{K}_\text{s}}}\xspace}

\newcommand{\Lmat}{\ensuremath{{\mathbf{L}}}\xspace}

\newcommand{\Cmats}{\ensuremath{{\mathbf{C}_\text{s}}}\xspace}

\newcommand{\nsimu}{\ensuremath{n_\text{sim}}\xspace}

\usepackage[colorinlistoftodos]{todonotes}

\begin{document}

\mainmatter  

\title{Continuous Optimization Benchmarks by Simulation\thanks{The final authenticated version is available online at \url{https://doi.org/10.1007/978-3-030-58112-1_19}}}


%
\author{Martin Zaefferer 
\and 
Frederik Rehbach} 
\authorrunning{M. Zaefferer and F. Rehbach}

\institute{Institute for Data Science, Engineering, and Analytics,\\
TH Köln, 51643 Gummersbach, Germany\\ 
\email{\{martin.zaefferer,frederik.rehbach\}@th-koeln.de}}

\maketitle

\begin{abstract}
Benchmark experiments are required to test, compare, tune, and understand optimization algorithms.
Ideally, benchmark problems closely reflect real-world problem behavior.
Yet, real-world problems are not always readily available for benchmarking. For example, evaluation costs may be too high, or
resources are unavailable (e.g., software or equipment).
As a solution, data from previous evaluations can be used to train surrogate models which
are then used for benchmarking. The goal is to generate test functions on which
the performance of an algorithm is similar to that on the real-world objective function.
However, predictions from data-driven models tend to be smoother than the ground-truth from which the training data is derived.
This is especially problematic when the training data becomes sparse.
The resulting benchmarks may not reflect the landscape features of the ground-truth, are too easy, and may lead to biased conclusions.

To resolve this, we use simulation of Gaussian processes instead of estimation (or prediction). 
This retains the covariance properties estimated during model training.
While previous research suggested a decomposition-based approach for 
a small-scale, discrete problem, we show that the spectral simulation method enables
simulation for continuous optimization problems.
In a set of experiments\footnote{
Reproducible code and a complete set of the presented figures is provided at \url{https://github.com/martinzaefferer/zaef20b}.
For easily accessible interfaces and demonstrations see \url{https://github.com/martinzaefferer/COBBS}.
}
with an artificial ground-truth, 
we demonstrate that this yields more accurate benchmarks
than simply predicting with the Gaussian process model.

\end{abstract}

\section{Introduction}
For the design and development of optimization algorithms, benchmarks are indispensable. 
Benchmarks are required to test hypotheses about algorithm behavior, 
to understand the impact of algorithm parameters, 
to tune those parameters,
or to compare algorithms with each other.
Multiple benchmarking frameworks exist, with BBOB/COCO being a prominent example~\cite{Hans16a,Hans19a}.

One issue of benchmarks is their relevance to real-world problems.
The employed test functions may be of an artificial nature, yet should reflect the behavior of algorithms on real-world problems. 
An algorithm's performance on test functions and real-world problems should be similar.
Yet, real-world problems may not be available in terms of functions, but only as data (i.e., observations from previous experiments). 
This can be due to real objective function evaluations being too costly or not accessible (in terms of software or equipment).

In those cases, using a data-driven approach may be a viable alternative: surrogate models can be trained and
subsequently used to benchmark algorithms.
The intent is not to replace artificial benchmarks such as BBOB (which have their own advantages), 
but rather to augment them with problems that have a closer connection to real-world problems.
This approach has been considered in previous investigations
\cite{Rudolph2009,Preuss2010,Bartz-Beielstein2015,Flasch2015b,Fischbach2016a,Dang2017}.
Additionally, recent benchmark suites offer access to real-world problems,
e.g., the Computational Fluid Dynamics (CFD) test problem suite~\cite{Daniels2018a} 
and the Games Benchmark for Evolutionary Algorithms (GBEA)~\cite{Volz2019a}.
Notably, the authors of the GBEA accept data provided by other researchers as a basis for surrogate model-based benchmarking\footnote{See the GBEA website, at \url{http://www.gm.fh-koeln.de/~naujoks/gbea/gamesbench_doc.html\#subdata}. Accessed on 2020-08-03.}.

As pointed out by Zaefferer et al.~\cite{Zaefferer2017a}, surrogate model-based benchmarks face a crucial issue:
the employed machine learning models may smoothen the training data, especially if the training data is sparse. 
Hence, these models are prone to produce optimization problems that lack the ruggedness and difficulty of the underlying real-world problems.
Thus, algorithm performances may be overrated, and comparisons become biased.
Focusing on a discrete optimization problem from the field of computational 
biology, Zaefferer et al. proposed to address this issue via simulation with Gaussian Process Regression (GPR).
In contrast to estimation (or prediction) with GPR, simulation 
may provide a more realistic assessment of an algorithm's behavior.
The response of the simulation retains the covariance properties determined by the model~\cite{Lantuejoul2002}.

The decom\-position-based simulation approach used by Zaefferer et al. relies on the selection of a
set of simulation samples~\cite{Zaefferer2017a}. 
The simulation is evaluated at these sample locations.
The simulation samples are distinct from and less sparse than the observed training samples.
They are not restricted by evaluation costs.
Still, using a very large number of simulation samples can quickly become computationally infeasible.
In small discrete search spaces, all samples in the search space can be simulated.
In larger search spaces, the simulation has to be interpolated between the simulation samples.
The interpolation step might again introduce undesirable smoothness.
Thus, decomposition-based simulation may work well for (small-scale) combinatorial optimization problems.
Conversely, it is not suited for continuous benchmarks.
Hence, our research questions are:
\begin{compactenum}[Q1]
\item How can simulation with GPR models be used to generate benchmarks for
continuous optimization?
\item Do simulation-based benchmarks provide better results than estimation-based benchmarks, for continuous optimization?
\end{compactenum}
For Q1, we investigate the spectral method for GPR-simulation~\cite{Cressie1993}.
The required background on GPR, 
estimation, and simulation is given in \cref{sec:gpr}.
Then, we describe a benchmark experiment to answer Q2 in \cref{sec:exp}, and the results in \cref{sec:res}. 
The employed code is made available.
We discuss critical issues of GPR and simulation in \cref{sec:modelissues}.
\Cref{sec:con} concludes the paper with a summary and outlook.

\section{Gaussian Processes Regression}\label{sec:gpr}
In the following, we assume that we deal with an objective function $\f(\vecx)$,
which is expensive to evaluate or has otherwise limited availability.
Here, $\vecx \in \RR^n$ are the variables of the optimization problem. 
Respectively, we have to learn models that regress data sets with $m$ training samples 
$\Xmat=\{ \vecx_1, \ldots, \vecx_m \}$, 
and the corresponding observations
$\vecy \in \RR^m$, 
with 
$y_j=\f(\vecx_j),$ 
and 
$ j=1,\ldots,m$.

\subsection{Gaussian Process Regression}
GPR (also known as Kriging) 
assumes that the training data $\Xmat$, $\vecy$ is
sampled from a stochastic process of Gaussian distribution.
Internally, it interprets data based on their correlations. 
These correlations are determined by a kernel $\kf(\vecx,\vecx')$.
A frequently chosen kernel is 
\begin{equation}
\label{eq:gausskern}
\kf(\mathbf{x},\mathbf{x}')=\exp\left(\sum_{i=1}^n -\theta_i |x_i-x'_i|^{2}\right).
\end{equation}
Here, $\theta_i \in \mathbb{R}$ is a parameter that is usually 
determined by Maximum Likelihood Estimation (MLE).
In the following, we assume that the model has already been trained via MLE,
based on the data $\Xmat$, $\vecy$.
The kernel $\kf(\vecx,\vecx')$ yields the correlation matrix $\Kmat$,
which collects all pairwise correlations of the training samples $\Xmat$.
The vector of correlations between each training sample $\vecx_j$, 
and a single, new sample $\vecx$ is denoted by $\veck$.
Further details on GPR, including model training by MLE, are given by Forrester et al.~\cite{Forrester2008a}.

In the context of GPR, the term \textit{estimation} denotes the prediction of the model at some unknown, new location.
It is performed with the predictor
\begin{equation}
\label{eq:pred}
\hat{y}(\vecx)=\hat{\mu}+\veck^T \Kmat^{-1} (\vecy-\bm{1}\hat{\mu}).
\end{equation}
Here, \evec is a vector of ones and the parameter $\hat{\mu}$ is determined by MLE.
Estimation intends to give an accurate response value at a single location \vecx.

\subsection{Simulation by Decomposition}
Conversely to estimation, \textit{simulation} intends to reproduce the covariance structure of a set of samples as accurately as possible~\cite{Journel1978,Cressie1993}.
Intuitively, this is exactly what we require for the generation of optimization benchmarks:
We are interested in the topology of the landscape (here: captured by the covariance structure), rather than accurate predictions of isolated function values~\cite{Preuss2010}.

One approach towards simulation is based on the decomposition of a covariance matrix $\Cmats$~\cite{Cressie1993}.
This matrix is computed for a set of $\nsimu$ simulation samples $\Xs =\{ \vecx_1, ... ,{\vecx}_{\nsimu} \}$,
with $\vecx_t \in \RR^n$ and $t=1,...,\nsimu$.
Here, \nsimu is usually much larger than the number of training samples $m$.
Using \cref{eq:gausskern}, \Xs  yields the correlation matrix $\Kmats$ of all simulation samples,
and the respective covariance matrix is $\Cmats=\hat{\sigma}^2 \Kmats$.
Here, $\hat{\sigma}^2$ is a model parameter (determined by MLE).
Decomposition can, e.g., be performed with the Cholesky decomposition
$\Cmats=\Lmat \Lmat^T$.
This yields the \textit{unconditionally} simulated values
$
\yvechs=\evec \hatmu +\Lmat \epsvec,
$
where \epsvec is a vector of independent normal-distributed random samples, $\epsilon_i \sim N(0,1)$.
In this context, `unconditional' means that the simulation reproduces only the covariance structure, but not the observed values \yvec. 
Additional steps are required for conditioning, so that the observed values are reproduced, too~\cite{Cressie1993}.

Obviously, the simulation only produces a discrete number of values \yvechs, at specific
locations \Xs.
Initially, we do not  know the locations where our optimization algorithms will attempt to evaluate the test function.
Hence, subsequent evaluations at arbitrary locations rely on interpolation. 
The predictor from \cref{eq:pred} can be used,
replacing all values linked to the training data with the respective values
from the simulation (\Xs, \yvechs, \Kmats instead of \Xmat, \yvec, \Kmat). The model parameters $\hat{\sigma}^2$, $\hat{\mu}$, and $\theta_i$ remain unchanged.

Unfortunately, this interpolation step is a critical weakness when applied to 
continuous optimization problems.
The locations \Xs have to be sufficiently dense, to avoid that the interpolation
introduces undesirable smoothness.
Yet, computational restrictions limit the density of \Xs.
Even a rather sparse grid of 20 samples in each dimension requires $m=20^n$ simulated samples. 
Then, $\Cmats$ is of dimension $20^n \times 20^n$, which is prohibitively large in terms of memory consumption for $n\geq4$.
Even a mildly multimodal function may easily require a much denser sample grid. 
This renders the approach infeasible for continuous optimization problems with anything but the lowest dimensionalities.

\subsection{Simulation by the Spectral Method}\label{sec:specSim}
Following up on~\cite{Zaefferer2017a}, we investigate a different simulation approach that is well suited for continuous optimization problems:
the spectral method~\cite{Cressie1993}.
This approach directly generates a function that can be evaluated at arbitrary locations, without interpolation.
It yields a superposition of cosine functions~\cite{Cressie1993},
\begin{equation*}
\fs (\vecx) = \hat{\sigma}  \sqrt{\frac{2}{N}} \sum_{v=1}^N \cos (\omegavec_v \cdot \vecx + \phi_v),
\end{equation*}
with $\phi_v$ being an i.i.d. uniform random sample from the interval $[ -\pi, \pi ]$.
The sampling of $\omegavec_v$ requires the spectral density function of the GPR model's kernel~\cite{Cressie1993,Lantuejoul2002}.
That is, $\omegavec_v \in \RR^n$ are i.i.d. random samples from a distribution with that same density.
For the kernel from \cref{eq:gausskern}, the respective distribution for the $i$-th dimension is the
normal distribution with zero mean and variance $2 \theta_i$.
A simulation conditioned on the training data can be generated with
$
\fsc (\vecx) = \fs (\vecx) + \hat{y}^*(\vecx)
$~\cite{Cressie1993},
where $\hat{y}^*(\vecx)=\hat{\mu}+\veck^T \Kmat^{-1} (\vecy_{sc}-\bm{1}\hat{\mu})$ is the predictor from \cref{eq:pred} with the training observations \vecy replaced by $\vecy_{sc}$, and $\hat{\mu}=0$. Here, $\vecy_{sc}$ are the unconditionally simulated values at the training samples, that is, $y_{\text{sc}_j} = \fs (\vecx_j)$.

\subsection{Simulation for Benchmarking}
To employ these simulations in a benchmarking context,
we roughly follow the approach by Zaefferer et al.~\cite{Zaefferer2017a}.
First, a data set is created by evaluating the true underlying problem (if not already available in the form
of historical data).
Then, a GPR model is trained with that data.
Afterwards, the spectral method is used to generate conditional or unconditional simulations.
These simulations are finally used as test functions for optimization algorithms.

\begin{figure}[tb]
  \centering
      \includegraphics[width=0.99\textwidth,page=1]{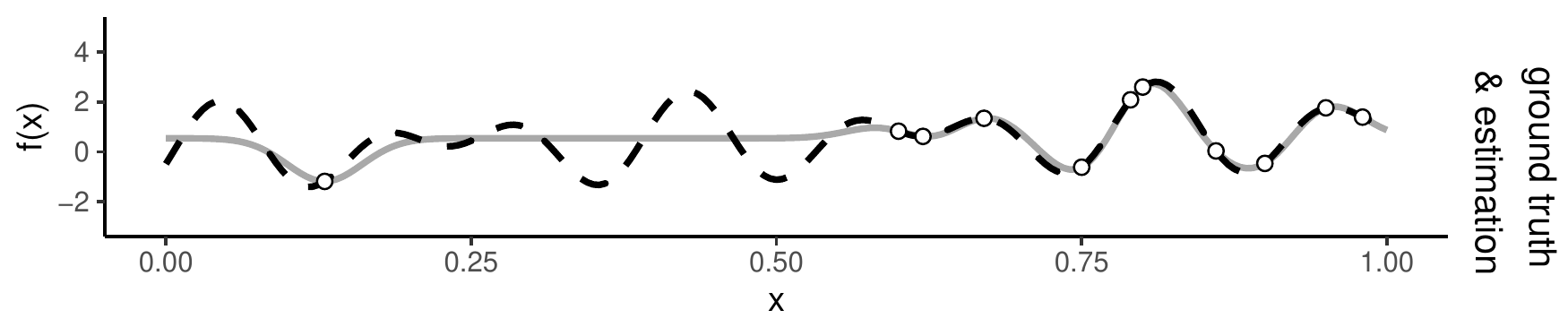}
      \includegraphics[width=0.99\textwidth,page=2]{illustrate}
      \includegraphics[width=0.99\textwidth,page=3]{illustrate}
  \caption{
  Top: Ground-truth $\f(x)$ (dashed line), training data (circles), and GPR model estimation (gray solid line).
  Middle: Three instances of an unconditional simulation (same model).
 Bottom: Three instances of a conditional simulation (same model).
The three different instances are generated by re-sampling of $\omegavec_v$ and $\phi_v$.
   }\label{fig:illu}
\end{figure}

Here, the advantage of simulation over estimation is the ability
to reproduce the topology of functions, rather than
predicting a single, isolated value. 
As an illustration, let us assume an example for $n=1$, where the ground-truth
is $\f(x)=\sin(33x) + \sin(49x - 0.5)  +x$. A GPR model is trained with
the samples $\Xmat = \{ 0.13,$ $0.6,$ $0.62,$ $0.67,$ $0.75,$ $0.79,$ $0.8,$ $0.86,$ $0.9,$ $0.95,$ $0.98\}$.
The resulting estimation, unconditional simulation, and conditional simulation 
of the GPR model are presented in \cref{fig:illu}. 
This example shows how estimation might be unsuited to reproduce an optimization algorithm's behavior.
In the sparsely sampled region, the GPR estimation is close to constant,
considerably reducing the number of local optima in the (estimated) search landscape.
The number of optima in the simulated search landscapes is considerably larger.



\section{Experimental Setup}\label{sec:exp}
In the following, we describe an experiment that compares test functions
produced by estimation and simulation with GPR.
\subsection{Selecting the Ground-Truth}
A set of objective functions is required as a ground-truth for our experiments. 
In practice, the ground-truth would be a real-world optimization problem.
Yet, a real-world case would limit the comparability, understandability, and the extent of
the experimental investigation.
We want to understand where and why our emulation deviates from the ground-truth.
This situation reflects the need for model-based benchmarks.

Hence, we chose a well-established artificial benchmark suite for optimization: the single-objective, noiseless BBOB suite
from the COCO framework~\cite{Hans16a,Hans19a}.
The BBOB suite allows us to compare in-detail how algorithms behave on the actual problem (ground-truth) and 
how they behave on an estimation or simulation with GPR.
Moreover, important landscape features of the BBOB suite are known (e.g., modality, symmetry/periodicity, separability), 
which enables us to understand and explain where GPR models fail.

The function set that we investigated is described in~\cite{Hansen2009a}.
This set consists of 24 unimodal and multimodal functions.  
For each function, 15 randomized instances are usually produced. We followed the same convention.
In addition, all test functions are scalable in terms of search space dimensionality $n$. 
We performed our experiments with $n= 2, 3, 5, 10, 20$.

\subsection{Generating the Training Data}\label{sec:trainingdata}
We generated training data by running an optimization algorithm on the original problem. 
The data observed during the optimization run was used to train the GPR model.
This imitates a common scenario that occurs in real-world applications:
Some algorithm has already been run on the real-world problem, and the data
from that preliminary experiment provides the basis for benchmarking. 
Moreover, this approach allows us to determine the behavior of the problem on a local and global scale. 
An optimization algorithm (especially, a population-based evolutionary algorithm) 
will explore the objective function globally as well as performing smaller, local search steps.

Specifically, we generated our training data as follows:
For each BBOB function $(1, ... , 24)$ and each function instance $(1,...,15)$,
we ran a variant of Differential Evolution (DE)~\cite{Storn1997} with $50 n$ function evaluations, and
a population size of $20 n$.
All evaluations were recorded.
We used the implementation from the DEoptim R-package, with default configuration~\cite{Ardia2020a}.
This choice  is arbitrary. Other population-based algorithms would
be equally suited for our purposes.

\subsection{Generating the Model}
We selected the $50n$ data samples provided by the DE runs.
Based on that data, we trained a GPR model, using the SPOT R-package~\cite{SPOTv20200429}.
Three non-default parameters of the model were specified with:
\verb|useLambda=FALSE| (no nugget effect, see also~\cite{Forrester2008a}),
\verb|thetaLower=1e-6| (lower bound on $\theta_i$), and
\verb|thetaUpper=1e12| (upper bound on $\theta_i$).
For the spectral simulation, we used $N=100 n$ cosine functions.
We only created conditional simulations, to reflect each BBOB instance as closely as possible. 
Scenarios where an unconditional simulation
is preferable have been discussed by Zaefferer et al.~\cite{Zaefferer2017a}.

\subsection{Testing the Algorithms}
We tested three algorithms:
\begin{itemize}
\item DE: As a global search strategy, we selected DE. We tested the same DE variant as mentioned in \cref{sec:trainingdata}, but with a population size of $10 n$ and a different random number generator seed for initialization. All other algorithm parameters remained at default values.
\item NM: As a classical local search strategy, the Nelder-Mead (NM) simplex algorithm was selected~\cite{Nelder1965}. We employed the implementation from the R-package nloptr~\cite{nloptrv1.2.1}.
All algorithm parameters remained at default values.
\item RS: We also selected a Random Search (RS) algorithm, which evaluates the objective function with i.i.d. samples from a uniform random distribution.
\end{itemize}

This selection was to some extent arbitrary.
The intent was not to investigate these specific algorithms.
Rather, we selected these algorithms to observe a range of different convergence behaviors.
Essentially, we made a selection that scales from very explorative (RS),
to balanced exploration/exploitation (DE), to very exploitative (NM).

All three algorithms receive the same test instances and initial random number generator seeds.
For each test instance, each algorithm uses $1000 n$ function evaluations.
Overall, each algorithm was run on each instance, function, and dimension of the BBOB test suite ($24\times15\times5= 1800$ runs, each run with $1000 n$ evaluations).
Additionally, each of these runs was repeated with an estimation-based test function,
and with a simulation-based test function.

\section{Results}\label{sec:res}
\subsection{Quality Measure}
Our aim is to measure how well algorithm performance is reproduced by the test functions.
One option would be to measure the error as the difference of observed values along the search path compared
to the ground-truth values at those same locations.
But this is problematic. Let us assume that the ground truth is $\f(x)=x^2$, and two test functions are $\f_\text{t1}(x)=(x-1)^2$ and $\f_\text{t2}(x)=0.5$, with $x \in [0,1]$.
Clearly, $\f_\text{t1}$ is a reasonable oracle for most algorithms' performance (e.g., in terms of convergence speed) while $\f_\text{t2}$ is not.
Yet, the mean squared error of $\f_\text{t1}$ would usually be larger than the error of $\f_\text{t2}$. 
The error on $\f_\text{t1}$ even increases when an algorithm approaches the optimum.

Hence, we measured the error on the performance curves.
For each algorithm run, the best observed function values after each objective function evaluation
were recorded (on all test instances, including estimation, simulation, and ground-truth).
In the following, this will be referred to as the performance of the algorithm.
The resulting performance values were scaled to values between zero and one, for each problem instance (i.e., each BBOB function, each BBOB instance, each dimension, and also separately for the ground-truth, estimation, and simulation variants).
This yielded what we term the scaled performance.
The error of the scaled performance was then calculated as the absolute deviation of the performance
on the model-based functions, compared to the performance on the ground-truth problem.
For example, let us assume that DE achieved a (scaled) function value on the ground-truth of 0.25 after 200 objective function evaluations.
But the same algorithm only achieved 0.34 on the estimation-based test function after 200 evaluations.
Then, the error of the estimation-based run is $|0.34-0.25|=0.09$ (after 200 evaluations).

\subsection{Observations}
\begin{figure}[tb]
  \centering
      \includegraphics[width=0.99\textwidth,page=1]{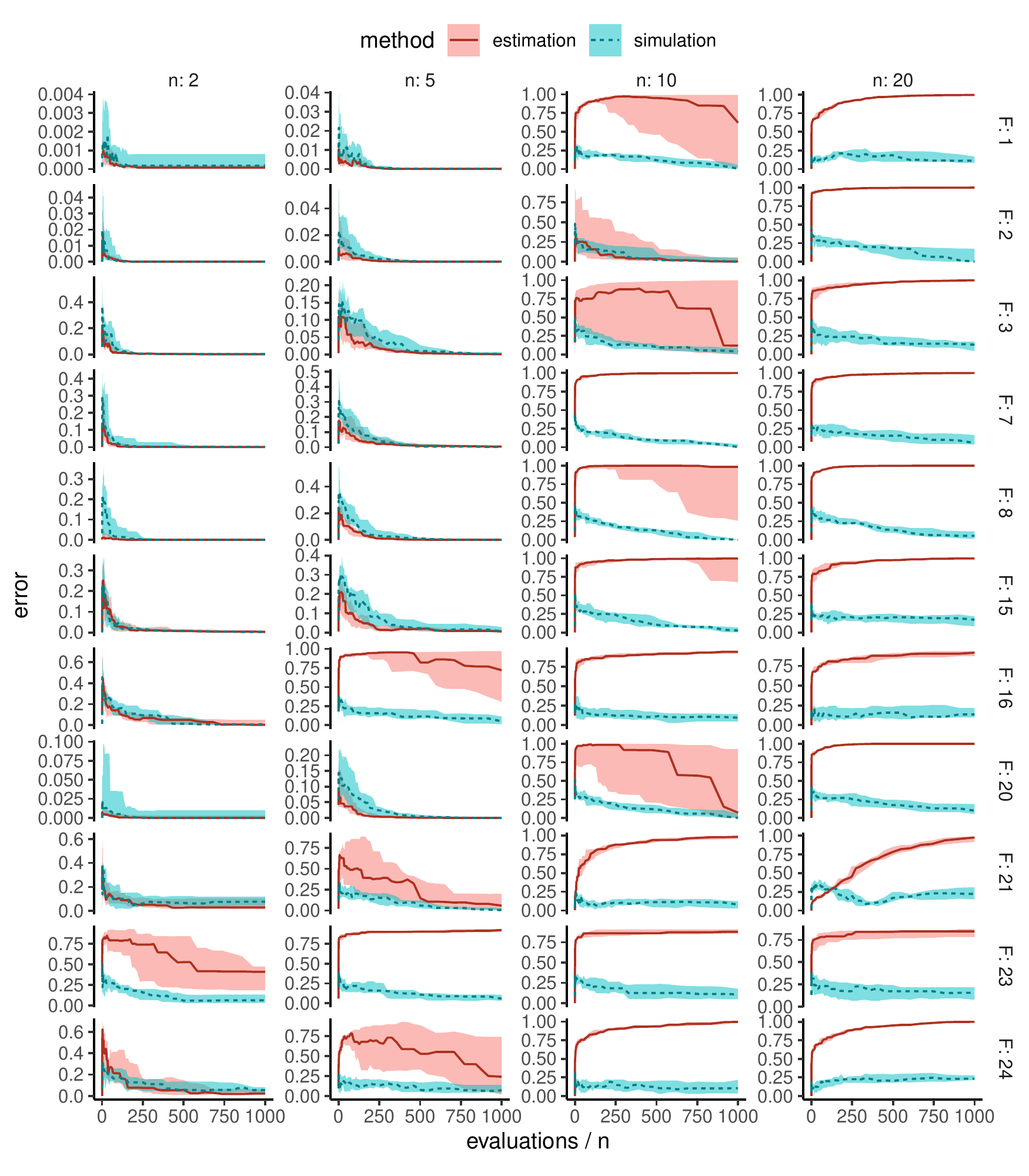}
  \caption{The error of algorithm performance with simulation- and estimation-based
  test functions. The curves are based on the performance of a DE run on the model-based
  test functions, compared against the performance values on the ground-truth, i.e., the respective BBOB functions.
  The labels on the right-hand side specify the respective IDs of the BBOB functions. Top-side labels indicate dimensionality. The lines indicate the median, the colored areas indicate the first and third quartile. These statistics are calculated over the 15 instances for each BBOB function.
  }\label{fig:selected}
\end{figure}

\begin{figure}[tb]
  \centering
      \includegraphics[width=0.99\textwidth,page=3]{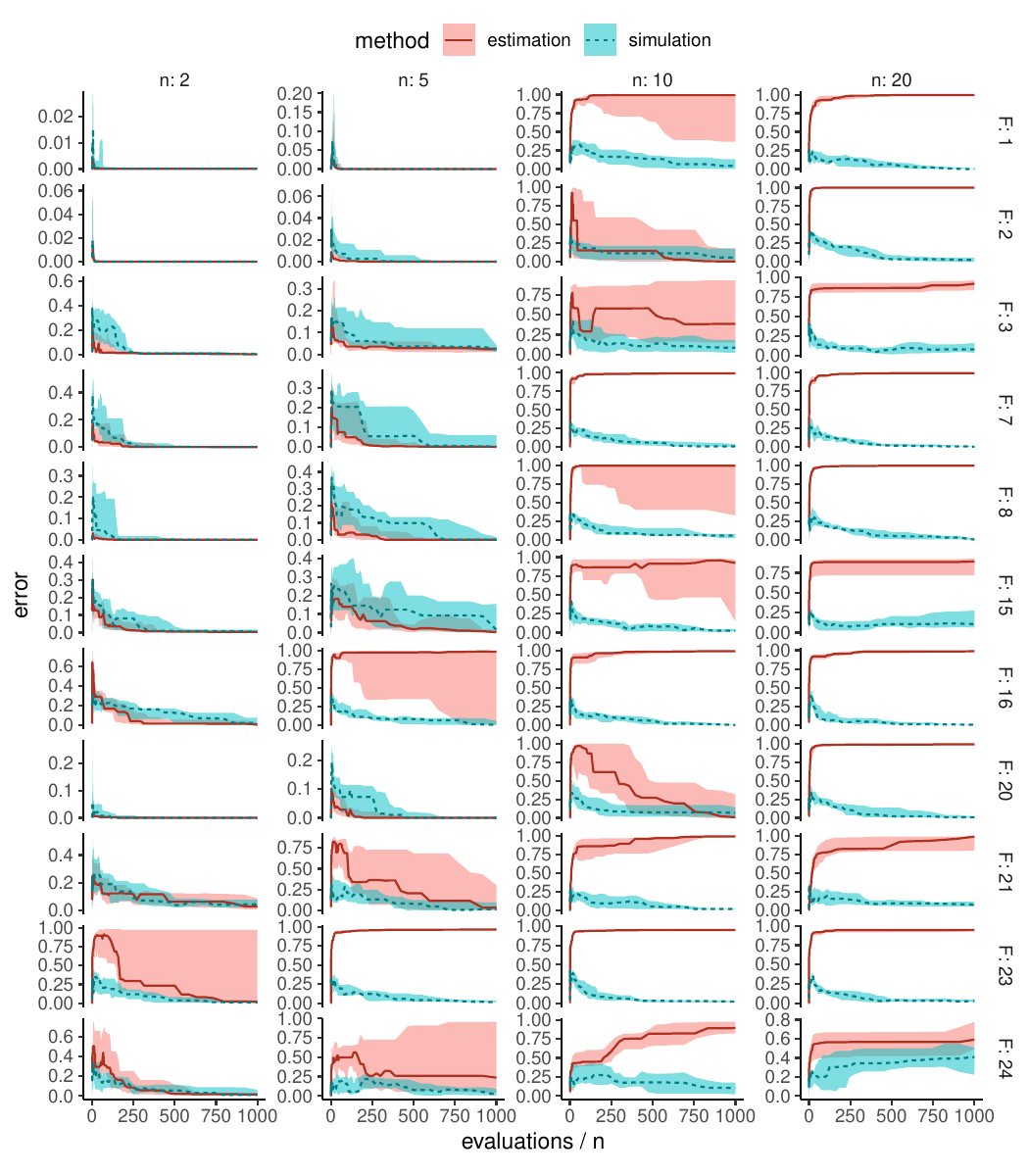}
  \caption{This is the same plot-type as presented in \cref{fig:selected}, but only for the performance of the NM algorithm (instead of DE).
  }\label{fig:selected2}
\end{figure}

In \cref{fig:selected,fig:selected2}, we show the resulting errors over run time for a subset of the 24 BBOB functions.
Due to space restrictions, we only show the error for the DE and NM algorithms. Similar patterns are visible
in the omitted curves for RS. We also omit the curves for $n=3$, which
closely mirror those for $n=2$.

For the simulation, we mostly observe decreasing errors or constant errors over time.
The decrease can be explained by the algorithms' tendency to find the best values for the respective problem instance later on in the run, regardless of the objective function. 
Earlier, the difference is usually larger.
For the estimation-based test functions, 
the error often increases, 
and sometimes decreases again during the later stages of a run.

When comparing estimation and simulation, the modality and the dimensionality $n$ are important.
For low-dimensional unimodal BBOB functions ($n=2, 3, 5$, and function IDs: 1, 2, 5-7, 10-14), the simulation yields larger errors than estimation.
In most of the multimodal cases, the simulation seems to perform equally well or better.
This can be explained: 
The larger activity of the simulation-based functions may occasionally introduce
additional optima (turning unimodal into multimodal problems). 
The estimation is more likely to reproduce the single valley of the ground-truth.
Conversely, the simulation excels for the multimodal cases because it does not remove optima by interpolation.
For higher-dimensional cases ($n=10,20$), this situation changes: the simulation produces lower
errors, regardless of modality. 
This is explained by the increasing sparseness of the training data, which in case of estimation will frequently lead to extremely poor search landscapes.
The estimation will mostly produce a constant value, with the exception of very small
areas close to the training data.

\begin{figure}[tb]
  \centering
      \includegraphics[width=0.99\textwidth,page=1]{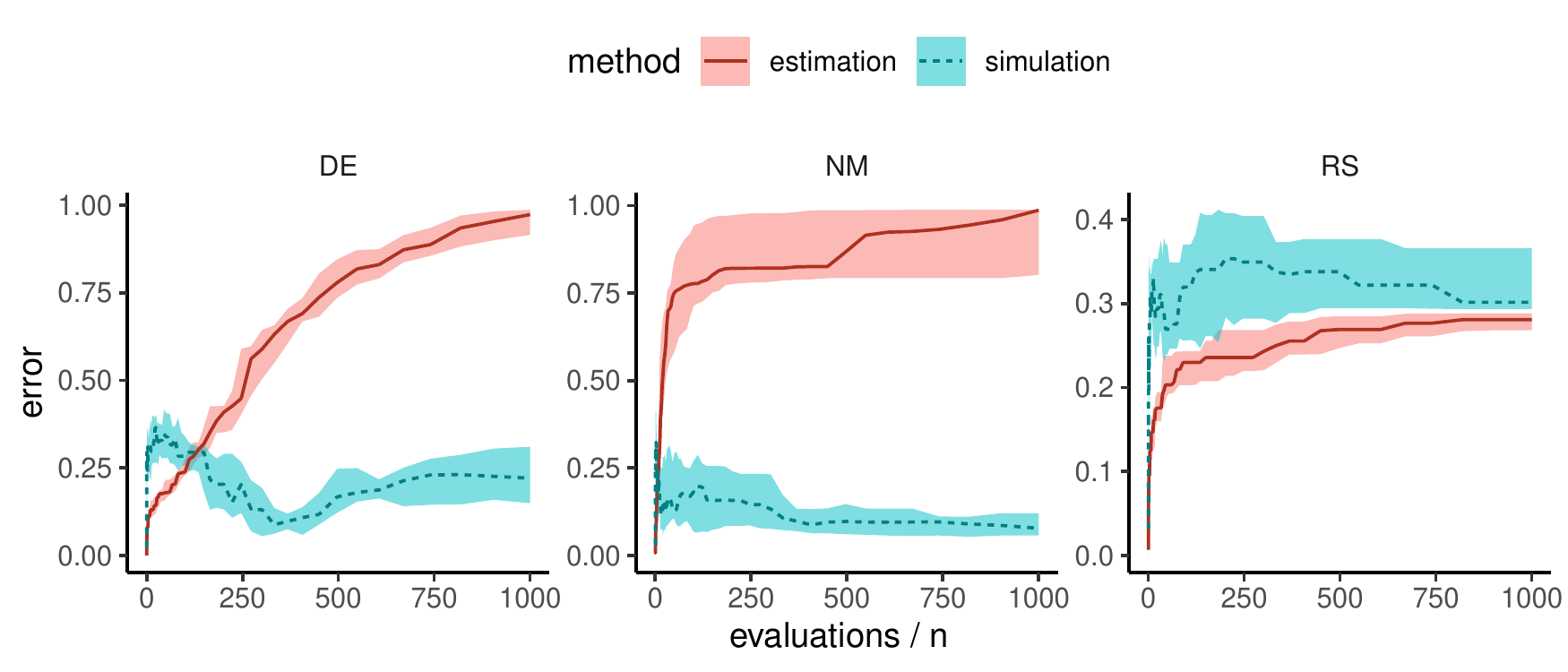}
  \caption{This is the same plot-type as presented in \cref{fig:selected}, but limited to BBOB function 21 and $n=20$. While \cref{fig:selected} only shows results for DE, this figure compares results for each tested algorithm (DE, NM, RS) for this specific function and dimensionality.}\label{fig:instance21}
\end{figure}
As noted earlier, results between DE, RS, and NM exhibit similar patterns.
There is an exception, where the results between DE, NM, and RS differ more strongly: BBOB function 21 and 22. 
Hence, \cref{fig:instance21} shows results for $n=20$ with function 21 (22 is nearly identical).
Here, each plot shows the error for a different algorithm.
Coincidentally, this includes the only case where estimation is performing considerably better
than simulation for large $n$ (with algorithm RS only).
The reason is not perfectly clear. 
One possibility is a particularly poor model quality.
BBOB functions 21 and 22 are both based on a mixture of Gaussian components.
Two aspects of this mixture are problematic for GPR:
Firstly, they exhibit a peculiar, localized non-stationarity. The activity of
the function may abruptly change direction, depending on the closest Gaussian
component.
Secondly, overlapping Gaussian components produce discontinuities.

\section{Discussion}\label{sec:modelissues}
The results show that the model-based test functions will occasionally deviate considerably from the ground-truth.
This has various reasons.
\begin{itemize}
\item \textbf{Dimensionality:} Clearly, all models are affected by the 
curse of dimensionality. 
With 10 or more variables, it becomes
increasingly difficult to learn the shape of the real function
with a limited number of training samples.
Necessarily, this limits how much we can achieve.
Despite its more robust performance, simulation also relies on a well-trained model.
\item \textbf{Continuity:} Our GPR model, or more specifically its kernel,
works best if the ground-truth is continuous, i.e., 
$\lim_{\vecx \to \vecx'} \f(\vecx) =\f(\vecx')$.
Else, model quality decreases.
One example is the step ellipsoidal function (BBOB function 7).
This weakness could be alleviated, if it is known a-priori: a more appropriate
kernel such as the exponential kernel
$\kf(\mathbf{x},\mathbf{x}')=$ $\exp(\sum_{i=1}^n -\theta_i |x_i-x'_i|)$ could be used.
However, this kernel may be less suited for the spectral simulation method~\cite{Cressie1993}.
\item \textbf{Non-stationarity:} Our GPR models assume
stationarity of the covariance structure in the data. 
Yet, some functions are obviously non-stationary.
For example, the BBOB variant of the Schwefel function (BBOB function 20) 
behaves entirely differently close to the search boundaries compared to the
optimal region (due to a penalty term).
In another way, the two Gallagher's Gaussian functions (BBOB functions 21, 22)
show a more localized type of non-stationarity. 
There, the activity of the function will change direction depending on the closest
Gaussian component.
Such functions are particularly difficult to model with common GPR models.
Non-stationary variants of GPR exist, and might be better suited. 
A good choice might be an approach based on clustering~\cite{Wang2017}.
Adapting the spectral method to that case is straight-forward.
The simulations from individual models (for each cluster) can be combined (locally) by a weighted sum.
\item \textbf{Regularity/Periodicity:} Several functions in the BBOB set
have some form of regular, symmetric, or periodic behavior. 
One classical example is the Rastrigin function (e.g., BBOB 3, 4 and 15).
While our models seemed to work well for these functions,
their regularity, symmetry or periodicity is not reproduced.
With the given models, this would require a much larger number of training samples.
If such behavior is important (and known a priori), a solution may be to choose
a new kernel that, e.g., is itself periodic.
This requires that the respective spectral measure of the new kernel is known,
to enable the spectral simulation method~\cite{Cressie1993}.
\item \textbf{Extremely fine-grained local structure:} Some functions,
such as Schaffer's F7 (BBOB function 17, 18), have an extremely fine-grained local
structure. 
This structure will quickly go beyond even a good model's
ability to reproduce accurately.
This will be true, even for fairly low-dimensional cases.
While there is no easy way out, our results at least suggest one compensation:
Many optimization algorithms will not notice such kind of fine-grained ruggedness.
For instance, a mutation operator might easily jump across these local bumps, and rather
follow the global structure of the function.
Hence, an accurate representation of such structures may not be 
that important in practice, depending on the tested algorithms.
\item \textbf{Number of samples:} The number of training data samples 
is one main driver of the complexity for GPR, affecting computational time 
and memory requirements. 
The mentioned cluster-GPR
approach is one remedy~\cite{Wang2017}.
\end{itemize}

\section{Conclusion}\label{sec:con}

Our first research question was:
\begin{compactenum}[Q1]
\item How can simulation with GPR models be used to generate benchmarks for
continuous optimization?
\end{compactenum}
As an answer, we use the spectral method for GPR simulation~\cite{Cressie1993}.
As this method results in a superposition of cosine functions,
it is well suited for continuous search spaces. Conversely, the previously used~\cite{Zaefferer2017a}
decomposition-based approach is infeasible due to computational issues.
Consecutively, we asked:
\begin{compactenum}[Q2]
\item Do simulation-based benchmarks provide better results than estimation-based benchmarks, for continuous optimization?
\end{compactenum}
Our experiments provide evidence that simulation-based benchmarks perform considerably better than estimation-based benchmarks.
Only for low-dimen\-sional ($n \leq 5$), unimodal problems did we observe an advantage for estimation.
In practice, if the modality (and dimensionality) of the objective function is known,
this may help to select the appropriate approach.
In a black-box case, the simulation approach seems to be the more promising choice.

For future research, it would be interesting to investigate how well these results translate
to non-stationary GPR models, as discussed in \cref{sec:modelissues}.
We also plan to investigate how parameters of the training data generation process
affect the generated test functions, perform tests with broader algorithm sets, and demonstrate
the approach with real-world applications. Finally, investigating other model types is of importance.
Approaches with weaker assumptions than GPR, such as Generative Adversarial Networks~\cite{Goodfellow2014}, may be
of special interest.

\bibliographystyle{abbrv}

\bibliography{zaefdiss}
\end{document}